\documentclass{article}

\usepackage{microtype}
\usepackage{graphicx}
\usepackage{subfigure}
\usepackage{booktabs} 

\usepackage{graphicx}
\usepackage{subfigure}  

\usepackage{hyperref}

\usepackage{amsmath}
\DeclareMathOperator*{\argmin}{arg\,min}

\usepackage[accepted]{icml2025}

\usepackage{amsmath}
\usepackage{amssymb}
\usepackage{mathtools}
\usepackage{amsthm}

\usepackage[capitalize,noabbrev]{cleveref}

\theoremstyle{plain}

\theoremstyle{definition}

\theoremstyle{remark}

\usepackage[textsize=tiny]{todonotes}

\icmltitlerunning{Direct Ascent Synthesis: Revealing Hidden Generative Capabilities in Discriminative Models}

\begin{document}

\twocolumn[

\icmltitle{Direct Ascent Synthesis: Revealing Hidden Generative Capabilities in Discriminative Models}

\begin{icmlauthorlist}
\icmlauthor{Stanislav Fort}{independent}
\icmlauthor{Jonathan Whitaker}{answer}
\end{icmlauthorlist}

\icmlaffiliation{independent}{Independent Researcher}
\icmlaffiliation{answer}{Answer.AI}

\icmlcorrespondingauthor{Stanislav Fort}{}

\vskip 0.3in
]

\printAffiliationsAndNotice{}  

\begin{abstract}
We demonstrate that discriminative models inherently contain powerful generative capabilities, challenging the fundamental distinction between discriminative and generative architectures. Our method, Direct Ascent Synthesis (DAS), reveals these latent capabilities through multi-resolution optimization of CLIP model representations. While traditional inversion attempts produce adversarial patterns, DAS achieves high-quality image synthesis by decomposing optimization across multiple spatial scales (1×1 to 224×224), requiring no additional training. This approach not only enables diverse applications -- from text-to-image generation to style transfer -- but maintains natural image statistics ($1/f^2$ spectrum) and guides the generation away from non-robust adversarial patterns. Our results demonstrate that standard discriminative models encode substantially richer generative knowledge than previously recognized, providing new perspectives on model interpretability and the relationship between adversarial examples and natural image synthesis.
\end{abstract}

\begin{figure}[t]
    \centering
    \includegraphics[width=1.0\linewidth]{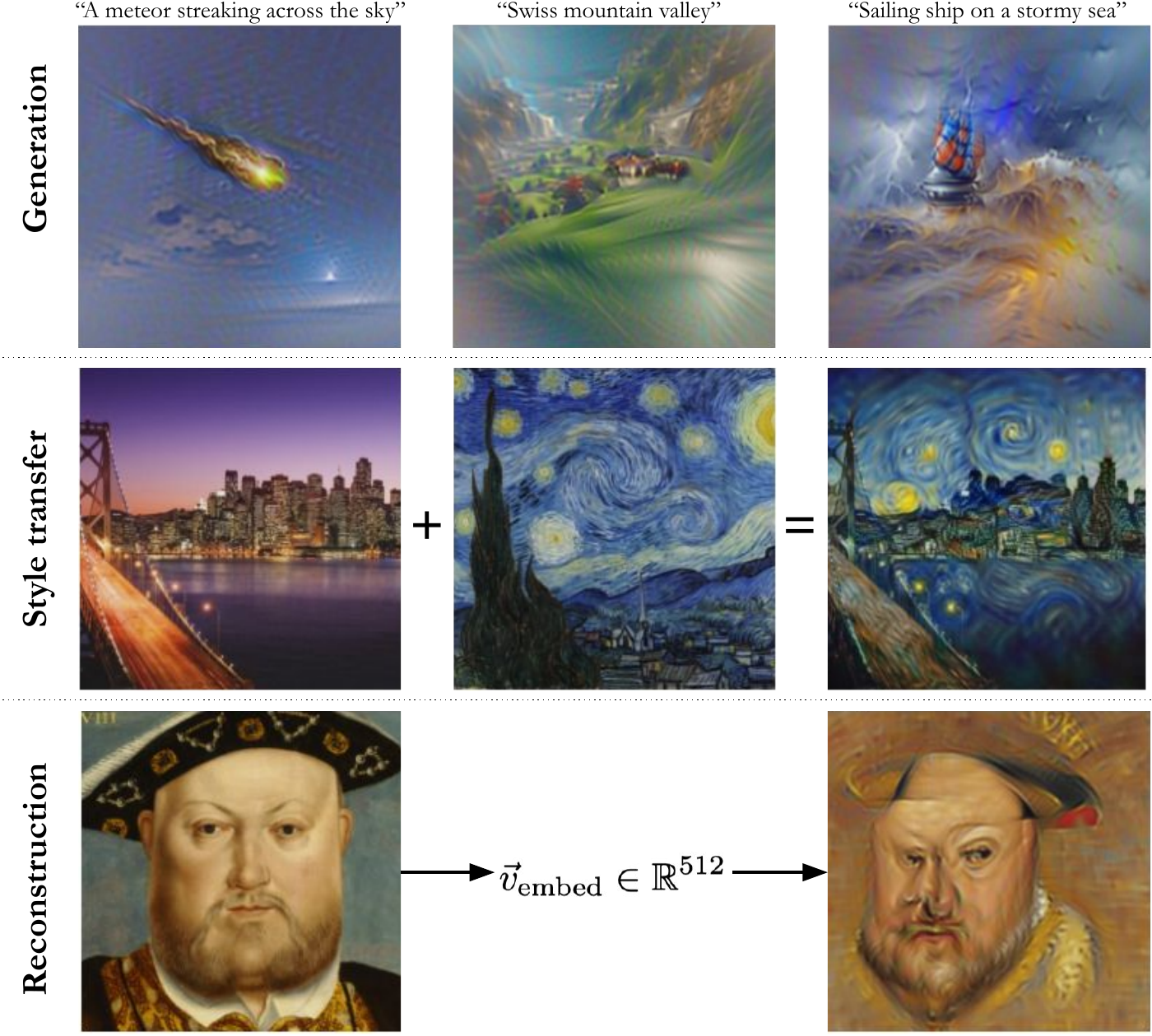}
    \caption{Direct Ascent Synthesis generates high-quality images by optimizing multi-resolution components to match CLIP embeddings, without any generative training. Unlike standard adversarial optimization that produces noise-like patterns, our approach reveals that pretrained discriminative models contain rich generative knowledge accessible through careful optimization. It can be used for a variety of image manipulations, such as style transfer and image reconstruction from a low-dimensional embedding.}
    \label{fig:subset-images}
\end{figure}

\section{Introduction}
\label{sec:intro}
\begin{figure*}[t]
    \centering
    \includegraphics[width=\textwidth]{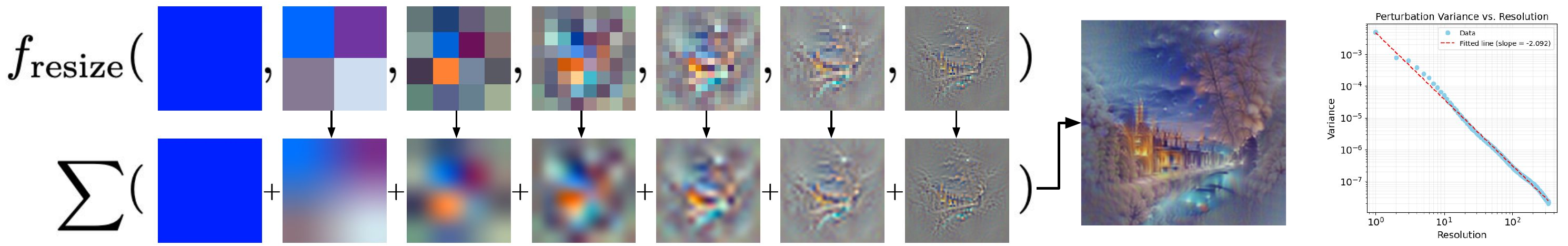}
    \caption{Multi-resolution decomposition enables training-free image synthesis. Left: An image is expressed as a sum of components at increasing resolutions, from $1\times1$ to $224\times224$. Middle: The components are optimized simultaneously to maximize CLIP embedding similarity with a target description, producing coherent images without generative training. Right: The power spectrum of generated images follows a $1/f^2$ distribution (slope $\approx-2$), characteristic of natural images. This demonstrates that our multi-resolution prior effectively guides optimization toward perceptually valid solutions.}
    \label{fig:multiresolution-sum}
\end{figure*}
Machine learning has traditionally relied on a fundamental dichotomy: discriminative models map inputs to semantic representations, while generative models synthesize data from learned latent spaces. This separation has driven remarkable progress, from GANs \cite{goodfellow2014generativeadversarialnetworks} to diffusion models \cite{ho2020denoisingdiffusionprobabilisticmodels, rombach2022highresolutionimagesynthesislatent}. However, these approaches require extensive training on large datasets, raising questions about whether such complex training procedures are fundamentally necessary for high-quality generation.

We challenge this dichotomy with Direct Ascent Synthesis (DAS), demonstrating that discriminative models implicitly encode rich generative knowledge accessible through careful optimization. The key challenge is that while discriminative models excel at mapping images to representations ($f: I \to v$), inverting this process ($f^{-1}: v \to I$) typically produces degenerate results. When optimizing an image $I^*$ to match a target representation $v$, the result often achieves near perfect mathematical alignment ($f(I^*) \approx v$) while appearing as meaningless noise to human observers. This phenomenon, first noted in the context of adversarial examples \cite{goodfellow2015explainingharnessingadversarialexamples}, reveals a fundamental tension between representation matching and perceptual quality. Trying to invert the latent representation back to a synthetic image was also the original motivation that led to the discovery of adversarial attacks.

Our key insight is that this degeneracy can be broken by decomposing the optimization across multiple resolutions. This multi-scale decomposition provides natural regularization that aligns with human visual priors, preventing degenerate high-frequency solutions while enabling explicit control over the generation process.

This simple approach produces high-quality, semantically meaningful images without any training, suggesting that pretrained discriminative models contain richer generative capabilities than previously recognized. DAS requires only seconds of computation on a single GPU for inference (and no compute for generative training at all), challenging assumptions about the necessity of extensive generative training. Beyond practical benefits, our work provides insights into the fluid boundary between discrimination and generation in deep neural networks—perhaps both types of models learn similar underlying representations, just accessed in different ways.

Our approach reveals a surprising connection between adversarial examples and image synthesis. The same optimization process that typically produces adversarial patterns can be redirected toward meaningful generation through appropriate regularization. This suggests that adversarial examples may not be a fundamental limitation of discriminative models, but rather a symptom of optimization that ignores natural image structure (this is explored in \cite{fort2024ensembleeverywheremultiscaleaggregation}).

We validate DAS through experiments on image generation, controlled modification, reconstruction, style transfer, and inpainting tasks. Our results demonstrate that combining discriminative representations with appropriate optimization priors enables high-quality synthesis without the computational and data requirements of traditional generative training.

\section{Related Work}
\begin{figure*}[t]
    \centering
    \includegraphics[width=\textwidth]{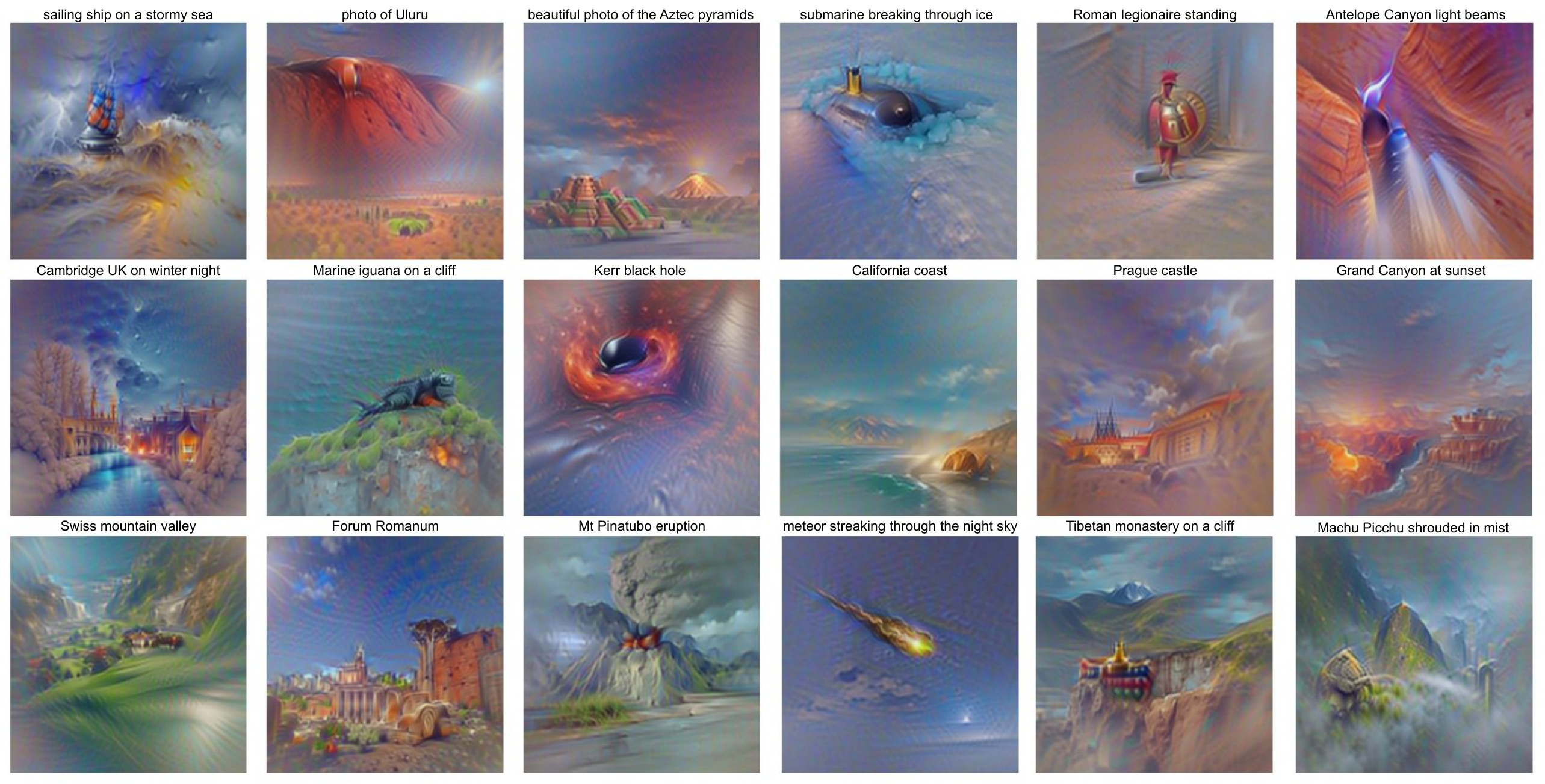}
    \caption{Diverse generations from Direct Ascent Synthesis across a range of concepts and styles. Results were obtained by optimizing against an ensemble of three CLIP models, with prompt augmentation to control image aesthetics: discouraging text generation (-0.3 × "Optical Character Recognition"), enhancing rendering quality (0.3 × "octane render, unreal engine, ray tracing, volumetric lighting"), and preventing image stacking (-0.3 × "multiple exposure").}
    \label{fig:all-images}
\end{figure*}
\subsection{The Evolution of Image Synthesis}
Image synthesis has traditionally followed two parallel tracks. The generative track progressed from VAEs \cite{kingma2022autoencodingvariationalbayes} and GANs \cite{goodfellow2014generativeadversarialnetworks} to diffusion models \cite{ho2020denoisingdiffusionprobabilisticmodels,rombach2022highresolutionimagesynthesislatent}, achieving remarkable quality through increasingly complex training. The discriminative track revealed rich internal representations through feature visualization \cite{yosinski2015understandingneuralnetworksdeep, nguyen2016synthesizingpreferredinputsneurons, olah2017feature} and adversarial examples \cite{goodfellow2015explainingharnessingadversarialexamples}, while models like CLIP \cite{radford2021learningtransferablevisualmodels} demonstrated that discriminative training can capture general visual concepts.

\subsection{Bridging Discrimination and Generation}
Several works have hinted at deeper connections between these approaches. Feature inversion methods \cite{mahendran2014understanding} showed that discriminative representations contain generative information, though with limited quality. Analysis of GAN discriminators \cite{bau2019seeinggangenerate} revealed latent spaces similar to generators, suggesting common representational principles. The success of optimization-based synthesis through techniques like deep image prior \cite{Ulyanov_2018_CVPR} and neural style transfer \cite{Gatys_2016_CVPR} demonstrated that careful optimization can sometimes replace explicit generative training.

More recently, the release of OpenAI's CLIP models \cite{radford2021learningtransferablevisualmodels} sparked a series of experiments in the open-source community that used CLIP similarity to a target text prompt to guide optimization in the latent space of various GAN generators. In particular, VQGAN-CLIP \cite{crowson2022vqgan} was used extensively for creative image generation and editing, and early practitioners quickly discovered the value of augmentations for improving and stabilizing the optimization process.

\subsection{Adversarial attacks on large models}
Despite early hopes to the contrary (especially due to scaling, e.g. \citet{dehghani2023scalingvisiontransformers22}), large models still suffer from adversarial examples. \citet{Fort2021CLIPadversarial, Fort2021CLIPadversarialstickers} shows that OpenAI CLIP models \citep{radford2021learningtransferablevisualmodels} can be fooled by small, easy-to-find, targeted, pixel-level modifications to the input image. Even very robust out-of-distribution detectors based on large scale pretrained models \citep{fort2021exploringlimitsoutofdistributiondetection} suffer from an equivalent brittleness under targeted attacks \citep{fort2022adversarialvulnerabilitypowerfulnear}. Transferable adversarial image attacks on proprietary models such as GPT-4, Claude and Gemini were first constructed in \citet{fort2024ensembleeverywheremultiscaleaggregation}. While there have been dedicated approaches improving adversarial robustness on small datasets (e.g. \citet{madry2019deeplearningmodelsresistant}), no solution has yet emerged at scale.

\subsection{The Role of Multi-Scale Processing}
The importance of multi-scale representations spans both classical and modern approaches \citep{Lindeberg01011994}, from Gaussian pyramids \cite{1095851} to recent architectures with explicit multi-scale processing \cite{fort2024ensembleeverywheremultiscaleaggregation}. This aligns with cognitive science findings that human visual processing operates across multiple spatial frequencies \cite{JEANTET2018123}. Our work builds directly on these insights by showing that multi-resolution optimization can bridge the gap between discriminative and generative processes.

The most directly related work is \citet{whitaker2024imstack}, which independently explored similar ideas of optimization-based image synthesis in their open source project.

\section{Multi-Resolution Optimization for Image Synthesis}

The foundation of our approach lies in understanding how natural images are structured across scales. Since the development of Gaussian and Laplacian pyramids \cite{1095851}, multi-resolution decomposition has been a powerful tool for analyzing images, revealing how information and statistics are organized across spatial frequencies \cite{d27961590ec04cb79e39942cf284a0ac}. We extend these classical insights to guide generative optimization in deep neural networks.

Our key innovation is reformulating image synthesis as simultaneous optimization across multiple scales:
\begin{equation}
    P(I) = \{I_r | r \in \rho\}, \text{ where } I_r = \text{rescale}_r(I)
\end{equation}
This decomposition serves several crucial purposes: 1) Provides natural regularization by enforcing consistency across scales, 2) Captures semantic information at appropriate resolutions, 3) Prevents degenerate high-frequency solutions characteristic of adversarial examples.

While traditional adversarial optimization often produces noise-like patterns by exploiting single-scale processing, our approach encourages consistency across the natural scale hierarchy of visual information. This aligns with both human visual processing---where different neural populations respond to features at different scales---and recent findings that multi-resolution processing improves neural network robustness \cite{fort2024ensembleeverywheremultiscaleaggregation}.

The optimization objective becomes:
\begin{equation}
    I^* = \argmin_{P_1,\ldots,P_R} \mathcal{L}\left(f\left(\sum_{r \in \rho} \mathrm{resize}_{224}(P_r)\right), v\right)
\end{equation}
where $P_r$ represents image components at resolution $r \times r$, and $\mathcal{L}$ measures representation similarity. This formulation automatically encourages solutions that respect the statistical structure of natural images while maintaining semantic coherence across scales.

This connection between scale-space consistency and natural image generation provides new insights into both adversarial robustness and generative modeling. The same principles that make representations robust against attack---consistency across scales and alignment with natural image statistics---also enable high-quality generation from discriminative models without additional training.

\section{Method}
\begin{figure}[t]
    \centering
    \includegraphics[width=1.0\linewidth]{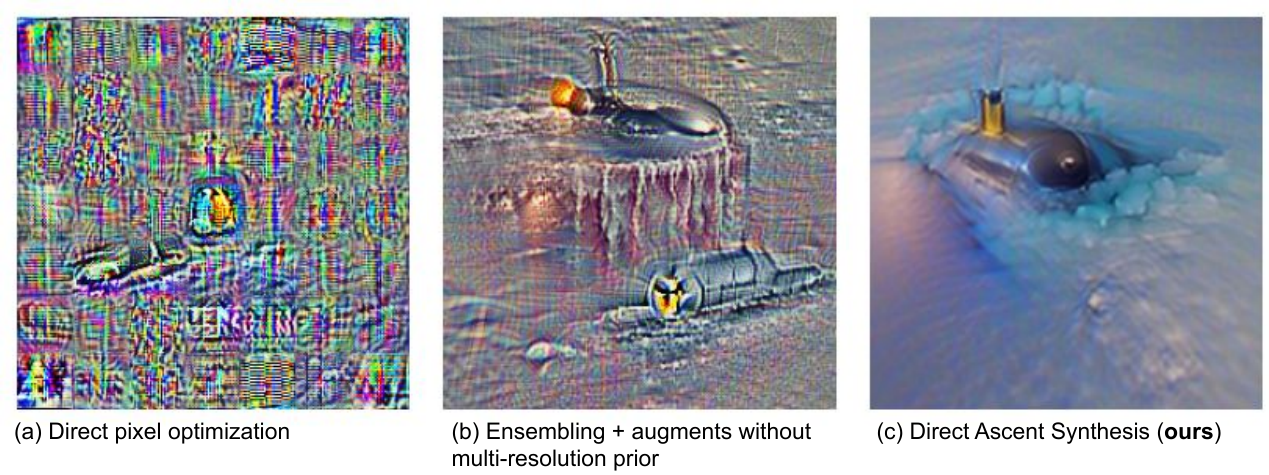}
    \caption{Ablation study demonstrating how different components of Direct Ascent Synthesis contribute to coherent image generation. Left: Direct pixel optimization yields adversarial patterns typical of model inversion attacks. Middle: Adding augmentations and model ensembling begins to impose structure but still lacks coherence. Right: Our complete approach with multi-resolution prior produces natural, interpretable images. This progression reveals how careful regularization can transform the degenerate solutions of model inversion into meaningful image synthesis.}
    \label{fig:role-of-resolution}
\end{figure}

\subsection{From Discrimination to Generation}

\begin{figure}[t]
    \centering
    \includegraphics[width=1.0\linewidth]{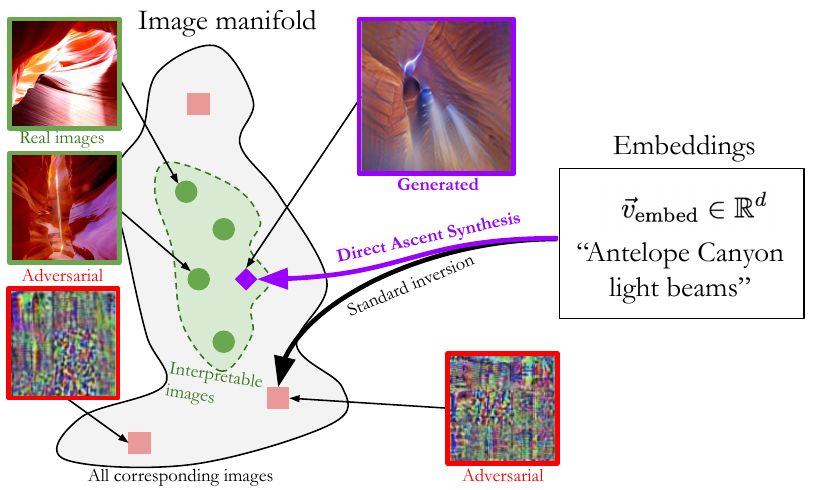}
    \caption{Mapping between images and embeddings. A region of all images corresponding to a \{text, image\} embedding contains interpretable images as well as noise-like adversarial patterns. Reconstructing an image from an embedding typically leads to such a degenerate noisy image. With Direct Ascent Synthesis, the reconstructed image lands among interpretable images within the manifold by default.}
    \label{fig:diagram}
\end{figure}

Every discriminative model contains within it the seeds of a generative model -- the challenge lies in accessing these capabilities effectively. Models like CLIP map images $I \in \mathbb{R}^{H \times W \times C}$ to embedding vectors $v \in \mathbb{R}^d$, learning rich representations that capture both semantic content and natural image structure. While the forward mapping ($f: I \to v$) is straightforward, the reverse mapping ($f^{-1}: v \to I$) has traditionally been seen as problematic due to its one-to-many nature and tendency to produce adversarial patterns (see Figure~\ref{fig:diagram} for a diagram capturing this). The dimensionality of such manifolds was studied in e.g. \cite{fort2022doesdeepneuralnetwork}.

Our key insight is that this perceived limitation is actually an opportunity: the space of possible inversions contains both natural images and adversarial patterns (as both are genuinely predictive of the embedding \citep{ilyas2019adversarialexamplesbugsfeatures}), and careful optimization can guide us toward the former. Given a target embedding $u$ (e.g., from a text description), we measure alignment through cosine similarity:

\begin{equation}
    \mathrm{score}(I) = \mathrm{cos}(f(I), u)
\end{equation}

\subsection{Multi-Resolution Optimization}
\begin{figure*}[!ht]
    \centering
    \includegraphics[width=\textwidth]{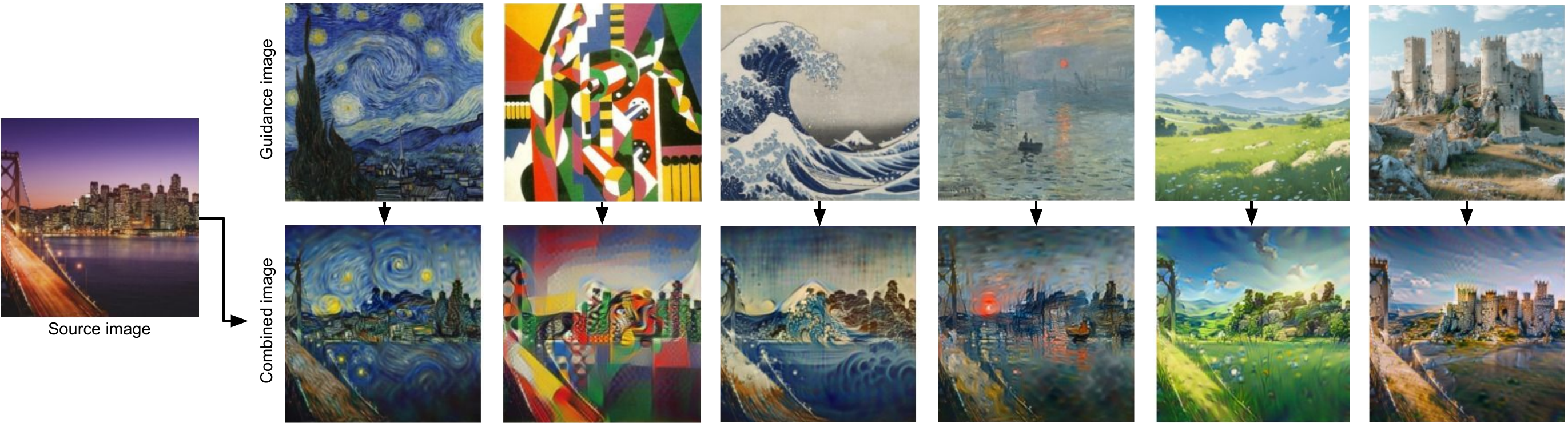}
    \caption{Direct Ascent Synthesis enables efficient neural style transfer without the artifacts common in pixel-space optimization. Starting from a source image and guidance image, we are able to effectively combine the two using DAS. This demonstrates that our multi-resolution framework naturally extends beyond CLIP-guided generation to other optimization-based image synthesis tasks.}
    \label{fig:das-style-transfer}
\end{figure*}

The critical innovation in DAS is decomposing the optimization across multiple scales -- a choice that proves surprisingly powerful in guiding solutions toward natural images. We break the degeneracy by decomposing the optimization across multiple scales. Instead of directly optimizing pixels, we express the image as a sum of resolution components:
\begin{equation}
    I = \frac{1}{2} + \frac{1}{2} \mathrm{tanh} \left(\sum_{r \in \rho} \mathrm{resize}_{224}(P_r)\right)
\end{equation}
where $P_r \in \mathbb{R}^{r \times r \times 3}$ represents the image component at resolution $r$, and $\rho$ spans from 1×1 to 224×224. The $\mathrm{tanh}$ transformation maps unbounded optimization values to valid pixel intensities while maintaining gradient flow.

The optimization objective becomes:
\begin{equation}
\sum_{i,j}\frac{\partial \mathrm{score}_i(\mathrm{augment}_j(I(P_1,\dots,P_{224})))}{\partial(P_1,\dots,P_{224})}
\end{equation}
where $i$ indexes multiple CLIP models and $j$ indexes augmentations. This formulation has several key properties: 1) Components are optimized simultaneously across all resolutions, 2) Gradients naturally distribute across scales based on their importance, 3) High-frequency adversarial patterns are suppressed by scale decomposition.

The resulting resolution components of a generated image are shown in Figure~\ref{fig:multiresolution-sum}, together with the power spectrum of the generated image, which follows a $1/f^2$ distribution (slope $\approx-2$), characteristic of natural images \citep{ruderman1994statistics, hyvarinen2009natural}. This demonstrates that our multi-resolution prior effectively guides optimization toward perceptually valid solutions.

\subsection{Implementation Details}
We employ several techniques to ensure stable and high-quality generation:

\paragraph{Augmentation.}
Two minimal augmentations prove crucial: random x-y shifts and pixel noise. These work in synergy with the multi-resolution prior -- neither is sufficient alone but together they enable robust generation. More complex augmentations might realistically lead to higher-quality generation.

\paragraph{Shift Handling.}
Rather than traditional padding approaches, we generate images at $(H+2s)\times(W+2s)$ resolution where $s$ is the maximum shift. This provides a natural buffer for shift augmentation, with the final image center-cropped to $H\times W$. Individual x-y shifts are guaranteed never to exceed the larger image, making sure padding is not necessary.

\paragraph{Model Ensemble.}
We average gradients across three CLIP models: OpenAI ViT-B/32 and two OpenCLIP \citep{Cherti_2023} ViT-B/32 variants trained on different datasets. This (marginally) improves generation quality, however, a single model is sufficient. We found, however, that some CLIP models were particularly bad at being turned into generators without any obvious reason why.

\begin{figure*}[!ht]
    \centering
    \subfigure[Volcanic eruption in Iceland]
    {\includegraphics[width=0.48\linewidth]{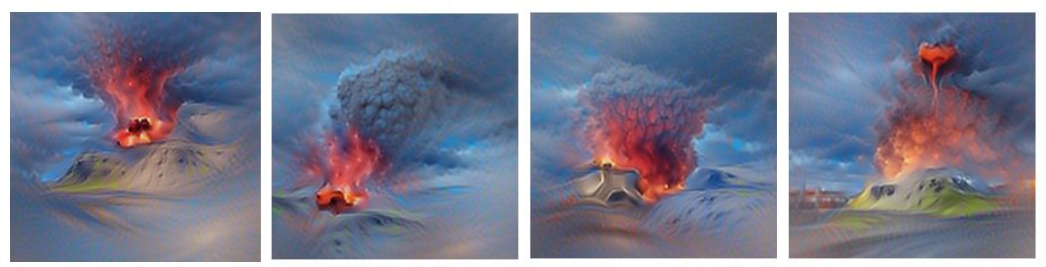}} \hfill
    \subfigure[Cambridge UK + winter night]
    {\includegraphics[width=0.48\linewidth]{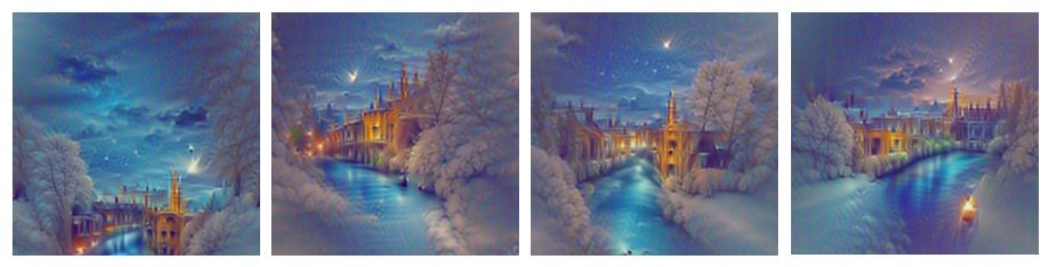}} \hfill
    \caption{Four independent generations of "a photo of a volcanic eruption in Iceland" on the left, and "a beautiful photo of Cambridge UK, detailed" with an additional prompt of "winter night". The generations used 3 CLIP models at once, and a corrective prompt of "Optical Character Recognition" with a weight of -0.6 to avoid text appearing in the images. The four samples demonstrate generation diversity.}
    \label{fig:versions}
\end{figure*}

\subsection{Extensions}
The framework naturally supports several useful extensions:

\paragraph{Multiple Target Vectors.}
Generation can be guided by multiple weighted targets: $\sum_i w_i \mathrm{score}(v, u_i)$. This enables fine control through prompt combinations (e.g., enhancing aesthetics with "volumetric lighting" while suppressing text with "Optical Character Recognition"). We use the latter to prevent CLIP from spelling out the semantic content of the desired generation.

\paragraph{Reference Images}
Target embeddings can come from either text or reference images ($f(I_\mathrm{ref}) = u$), enabling style transfer and reconstruction tasks. Despite CLIP's compression from 150,528 dimensions to 512, reconstruction often preserves both semantic content and style elements. See Figure~\ref{fig:das-style-transfer} for style transfer and Figure~\ref{fig:reconstruction} for reconstruction examples.

\section{Experiments and Analysis}
We evaluate Direct Ascent Synthesis through a comprehensive set of experiments designed to probe both its generative capabilities and its relationship to discriminative representations. Our analysis focuses on four key aspects of the method: generation consistency, controlled modification, reconstruction fidelity, and versatility across different applications.

The specific optimization details were kept as simple as possible for the sake of understanding the underlying generative capabilities of DAS. We optimized for 100 steps with Stochastic Gradient Descent \citep{Robbins_1951} at the learning rate of $2\times10^{-1}$. The added noise had a standard deviation of 0.2, the x-y plane jitter in the $\pm56$ range (implying a generation of a $336\times336$ from which we then center-cropped the $224\times224$ in the middle), and using 32 augmentations at once in a batch. We also used 3 CLIP models in an ensemble: OpenAI ViT-B/32 and two OpenCLIP ViT-B/32. All models we used are based on the Vision Transformer architecture \citep{dosovitskiy2021imageworth16x16words}, however, we have verified that non-ViT models work similarly well.

Figure~\ref{fig:all-images} shows 18 images generated using DAS from a diverse set of prompts.

\subsection{Generation Quality and Consistency}
A fundamental question for any generative method is whether it can consistently produce coherent results. Figure~\ref{fig:versions} demonstrates DAS's reliability across multiple runs on two challenging prompts: a dynamic natural phenomenon (volcanic eruption) and a complex architectural scene (Cambridge on a winter night). These examples were chosen specifically to test the method's ability to handle both natural and man-made structures, dynamic events, and specific lighting conditions.

Our analysis reveals three key properties of the generation process:

\begin{itemize}
    \item \textbf{Semantic Consistency:} Each set of generations maintains consistent high-level features while varying in specific details. For the volcanic scenes, this manifests as consistent plume structure and landscape integration. In the Cambridge scenes, we observe reliable architectural motifs and winter atmosphere, suggesting that the optimization reliably finds meaningful regions in CLIP's representation space.
    
    \item \textbf{Compositional Understanding:} The images demonstrate sophisticated composition without explicit training. The volcanic scenes balance foreground drama with environmental context, while the Cambridge scenes show an understanding of architectural perspective and nighttime illumination. This suggests that our multi-resolution optimization effectively accesses CLIP's learned understanding of scene structure.
    
    \item \textbf{Natural Variation:} The differences between runs exhibit variations characteristic of natural images—lighting changes, slight perspective shifts, and detail variations—rather than adversarial patterns. This indicates that our multi-resolution prior successfully constrains the optimization to the natural image manifold.
\end{itemize}

\subsection{Controlled Modification}
\begin{figure}[t]
    \centering
    \subfigure[Original]{\includegraphics[width=0.48\linewidth]{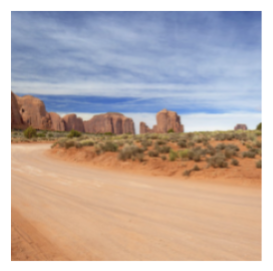}} \hfill
    \subfigure["snowy road"]{\includegraphics[width=0.48\linewidth]{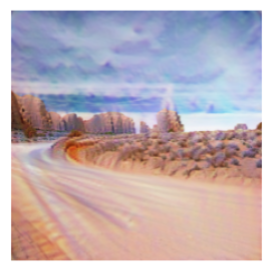}} \\
    \subfigure["a volcano eruption"]{\includegraphics[width=0.48\linewidth]{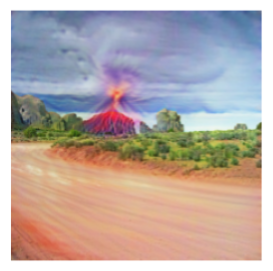}} \hfill
    \subfigure["asphalt road"]{\includegraphics[width=0.48\linewidth]{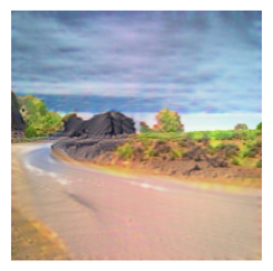}}
    \caption{Starting from the original image, we run generation towards the specified prompt. The modifications demonstrate both local changes (road surface, volcano in the background) and global scene transformations (winter and snow) while maintaining spatial coherence.}
    \label{fig:road-change}
\end{figure}

Figure~\ref{fig:road-change} explores DAS's capacity for targeted image modification, demonstrating both local adjustments (surface changes) and global transformations (environmental shifts). These experiments reveal several important capabilities:

\begin{itemize}
    \item \textbf{Structure Preservation:} Core scene geometry and spatial relationships persist across transformations, indicating that our optimization respects structural features encoded in CLIP's representation space.
    
    \item \textbf{Semantic Control:} The modifications show precise response to textual prompts while maintaining physical plausibility. Snow accumulates naturally on surfaces, the volcano emerges with appropriate atmospheric effects, and surface textures change coherently.
    
    \item \textbf{Multi-Scale Coordination:} New elements integrate seamlessly across different spatial scales. This is particularly evident in the volcano example, where both large-scale landscape changes and local atmospheric effects are coordinated.
\end{itemize}

\subsection{Embedding-Guided Reconstruction}
\begin{figure}[t]
    \centering
    \includegraphics[width=1.0\linewidth]{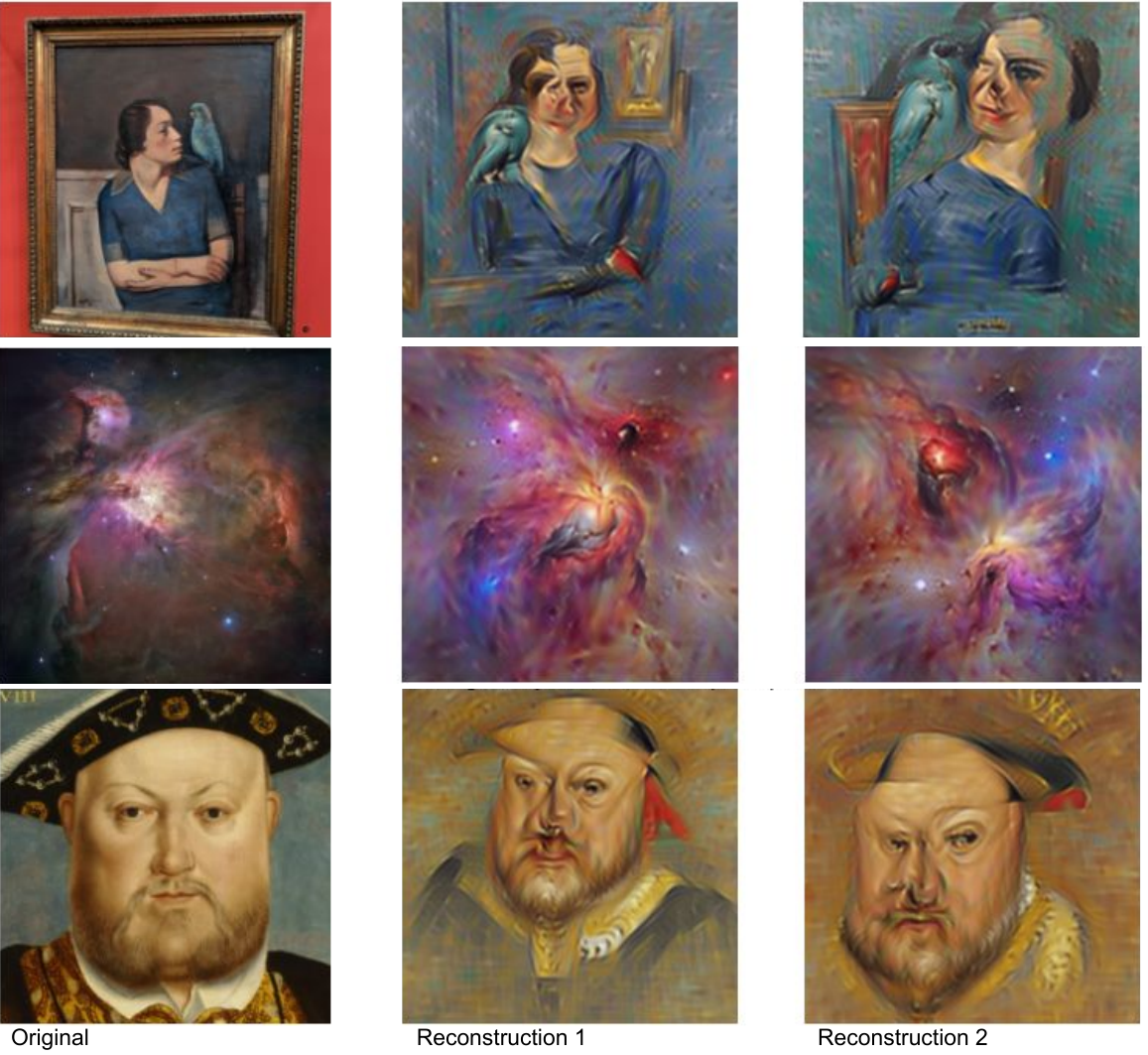}
    \caption{Reconstructing an image from its embedding. Instead of a text prompt, we used an embedded original image to guide the Direct Ascent Synthesis generation. Two resulting reconstructions are shown for each image, demonstrating consistent recovery of major semantic elements and style while allowing natural variations in specific details. Given the 300:1 dimensionality reduction from an image to an embedding, the recovery is impressive.}
    \label{fig:reconstruction}
\end{figure}

Image reconstruction from CLIP embeddings provides a particularly rigorous test of our method, as it requires recovering high-dimensional image structure from highly compressed representations (from 150,528 dimensions to just 512). Figure~\ref{fig:reconstruction} demonstrates that DAS can recover substantial semantic and stylistic information, with the reconstructions showing:

\begin{itemize}
    \item \textbf{Semantic Preservation:} Major scene elements and their relationships are consistently recovered
    \item \textbf{Style Retention:} Color schemes, lighting conditions, and artistic qualities transfer effectively
    \item \textbf{Compositional Fidelity:} Overall layout and spatial organization remain intact
    \item \textbf{Natural Variation:} Fine details vary while maintaining scene coherence
\end{itemize}

This performance is particularly noteworthy given that CLIP was never trained for reconstruction or compression tasks. Yet it can recover major aspects of an image from its 300:1 compressed embedding.

\subsection{Specialized Applications}
\begin{figure}[t]
    \centering
    \subfigure[Japan]{\includegraphics[width=0.24\linewidth]{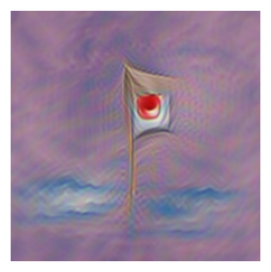}} \hfill
    \subfigure[Brazil]{\includegraphics[width=0.24\linewidth]{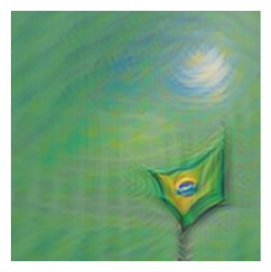}} \hfill
    \subfigure[UN]
    {\includegraphics[width=0.24\linewidth]{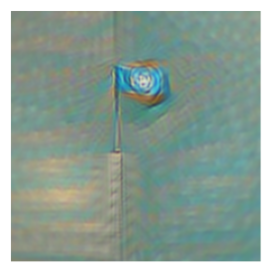}} \hfill
    \subfigure[Swiss]
    {\includegraphics[width=0.24\linewidth]{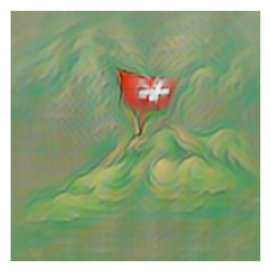}} \hfill
    \caption{Generating flags demonstrates DAS's ability to handle precise geometric patterns and symbolic elements. The prompts combine "the flag of [X]" with structural guidance (-0.3 × "Optical Character Recognition", 0.6 × "cohesive single subject", -0.3 × "multiple exposure").}
    \label{fig:flags}
\end{figure}

\begin{figure}[t]
    \centering
    \includegraphics[width=1.0\linewidth]{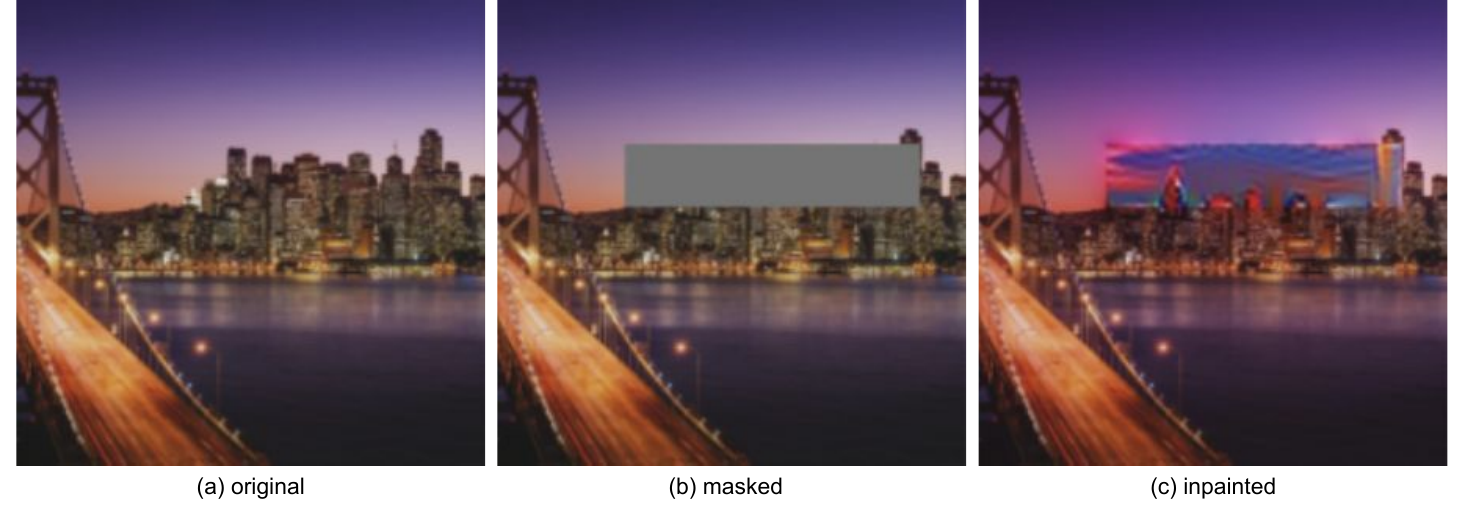}
    \caption{An example of inpainting using Direct Ascent Synthesis. The masked image was filled in using the prompt "a city skyline at night", demonstrating seamless integration of generated content with existing context.}
    \label{fig:inpaiting}
\end{figure}

To explore the versatility of our framework, we tested DAS on specialized generation tasks that typically require dedicated solutions. Figure~\ref{fig:flags} shows the generation of national flags, a task requiring precise geometric patterns and symbolic elements. The results demonstrate that DAS can handle both rigid geometric constraints and subtle style elements, like the precise proportions of the Swiss cross and the complex star pattern of the Brazilian flag. While the generation is far from perfect, the flags are clearly recognizable.

Figure~\ref{fig:inpaiting} showcases inpainting capabilities, where DAS must generate content that seamlessly integrates with existing image context. The successful completion of the city skyline demonstrates that our multi-resolution optimization naturally handles boundary conditions and structural continuity without additional constraints.

These specialized applications highlight a key advantage of our approach: a single optimization framework can address diverse synthesis tasks without task-specific training or architectural modifications. This versatility stems from the rich representational knowledge captured by CLIP combined with the natural structure-preserving properties of multi-resolution optimization.

\subsection{Style Transfer}
We can modify a starting image toward the embedding of a "style" image easily using DAS. This functions effectively as a natural version of style transfer. See Figure~\ref{fig:das-style-transfer} for details. The resulting generations respect the structure of the original image while copying the style and local content from the guiding image.

\begin{figure}[t]
    \centering
    \subfigure[Style]{\includegraphics[width=0.24\linewidth]{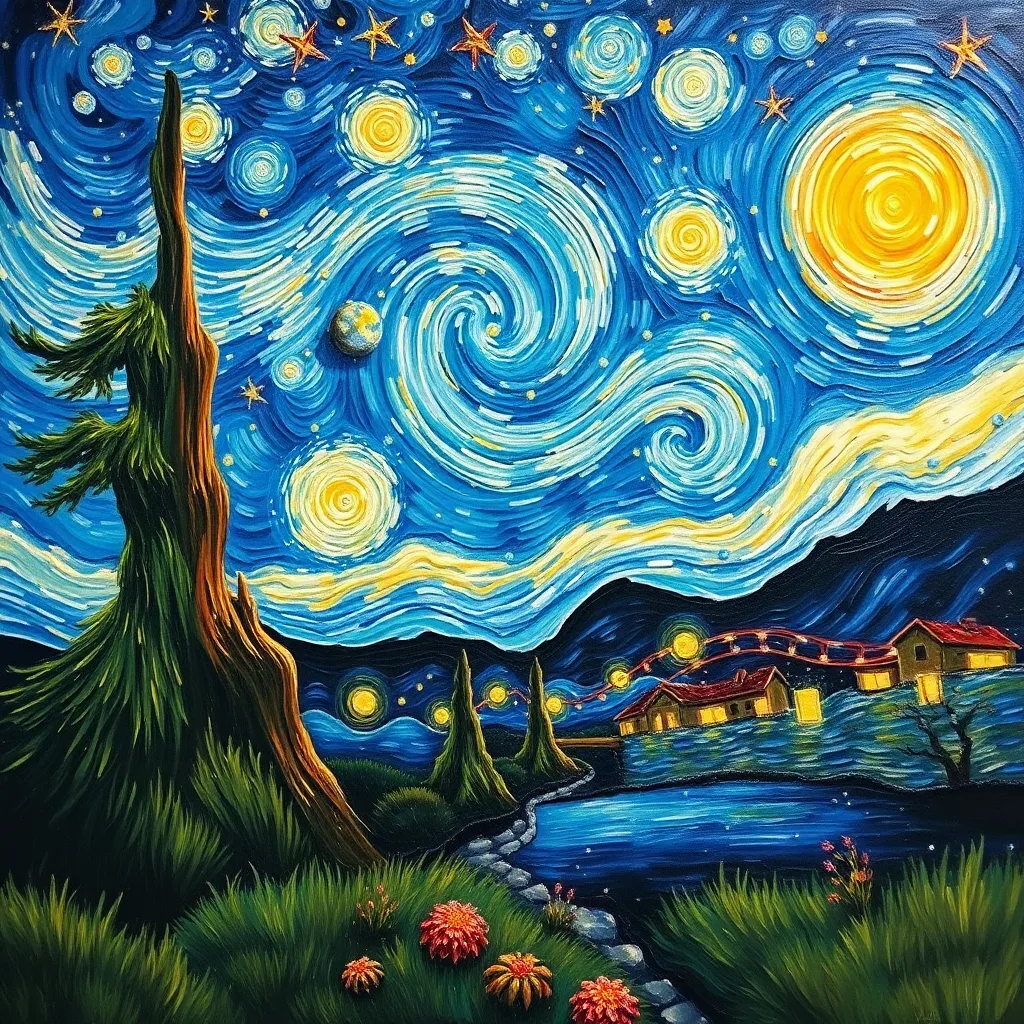}} \hfill
    \subfigure[Content]{\includegraphics[width=0.24\linewidth]{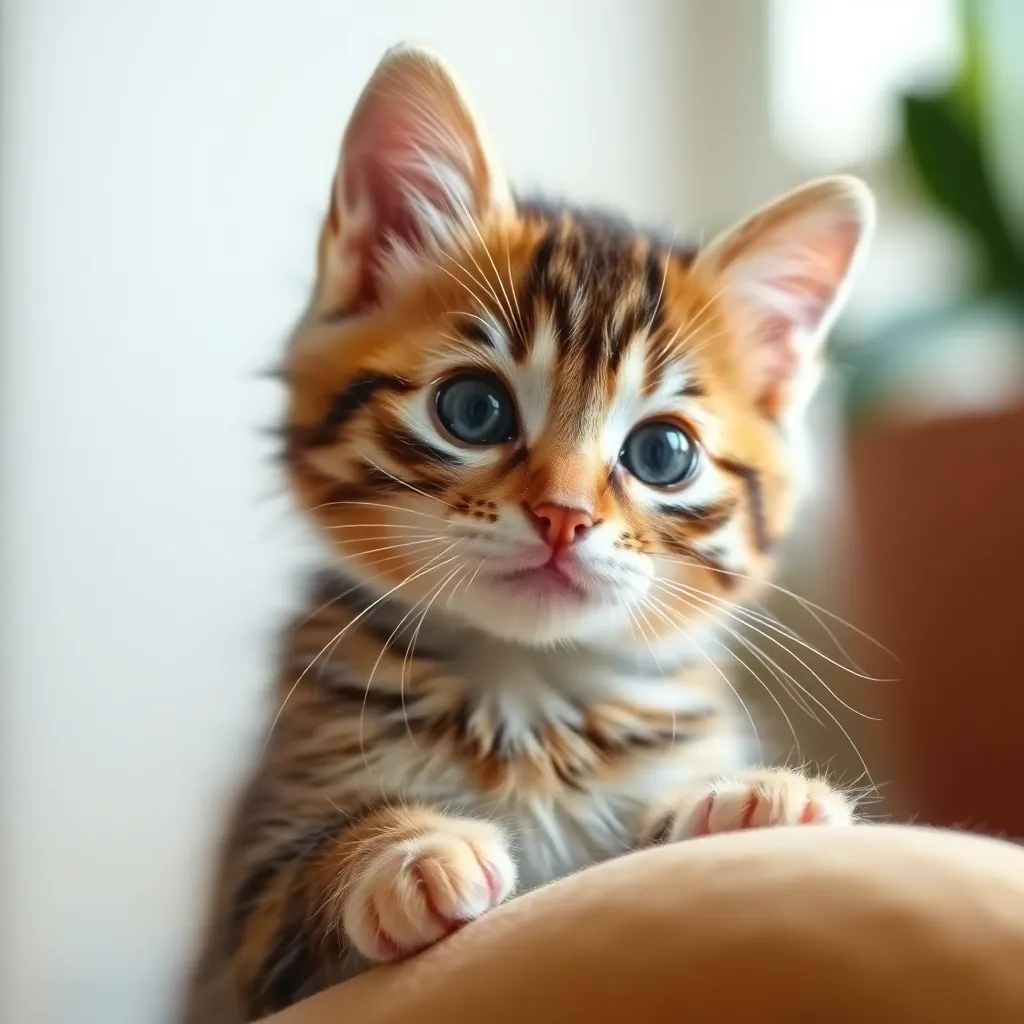}} \hfill
    \subfigure[Pixels]
    {\includegraphics[width=0.24\linewidth]{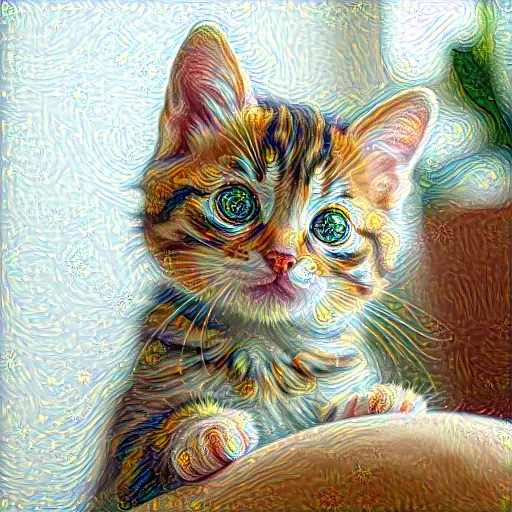}} \hfill
    \subfigure[DAS]
    {\includegraphics[width=0.24\linewidth]{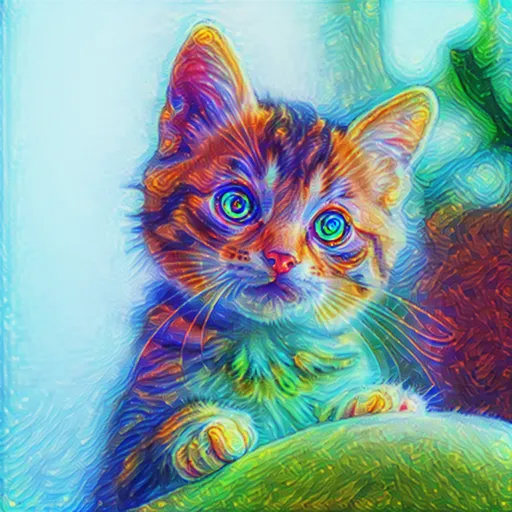}} \hfill
    \caption{Applying style transfer while optimizing raw pixels (c) vs DAS (d).}
    \label{fig:standard-style-transfer}
\end{figure}

Beyond CLIP, we find DAS a useful drop-in replacement for other techniques that rely on pixel-space optimization, such as the more traditional style transfer approaches. For example, following \cite{Gatys_2016_CVPR} we compare the results on raw pixels with those obtained by using our method and, while the comparison is somewhat subjective, find that the latter tends to produce more pleasing results with less high-frequency artifacts and in fewer steps. See Figure~\ref{fig:standard-style-transfer}.

\section{Discussion and Future Work}
Our results with Direct Ascent Synthesis reveal fundamental insights about the relationship between discrimination and generation in deep neural networks. By demonstrating that discriminative models contain rich generative knowledge that can be accessed through careful optimization, we challenge several conventional assumptions in the field.

\subsection{Theoretical Implications}
The success of DAS suggests that the traditional division between discriminative and generative models may be more fluid than previously thought. It appears that seeds of generation are hidden in all discriminative models, and we just needed a better way of eliciting them. Several key insights emerge:

\begin{itemize}
    \item \textbf{Representation Unification:} The ability of discriminative models to support high-quality generation suggests a fundamental unity in neural representations. Rather than learning strictly task-specific features, these models appear to capture a more complete understanding of visual structure that can support both discrimination and generation. This challenges the traditional view of separate representational requirements for these tasks.
    
    \item \textbf{Information Preservation:} Our results indicate that discriminative training, despite optimizing only for classification or similarity metrics, preserves much of the information needed for generation. This suggests that the process of learning to discriminate naturally encodes generative capabilities as a byproduct, pointing to deeper connections between these two aspects of visual processing.
    
    \item \textbf{Optimization vs. Architecture:} DAS demonstrates that the key to accessing generative capabilities may lie more in the optimization process than in network architecture. This suggests that the historical focus on architectural differences between discriminative and generative models may have overshadowed the importance of how we access and utilize their learned representations.
\end{itemize}

\subsection{Practical Implications}
Beyond theoretical insights, DAS has several important practical implications:

\begin{itemize}
    \item \textbf{Resource Efficiency:} By leveraging pretrained discriminative models, DAS enables image generation with significantly lower computational requirements than traditional generative approaches.
    
    \item \textbf{Architecture Design:} The success of multi-resolution optimization suggests new directions for neural architecture design that explicitly incorporate scale-space structure.
    
    \item \textbf{Model Reuse:} DAS demonstrates that existing discriminative models may have untapped capabilities that can be accessed through novel optimization strategies.
\end{itemize}

\subsection{Connections to Model Interpretability}

Our work with DAS has significant implications for model interpretability research. The ability to extract coherent generative capabilities from discriminative models suggests that standard interpretability approaches may overlook important model properties. While traditional interpretability tools focus on analyzing individual neurons or attention patterns, DAS reveals emergent capabilities that arise from the interaction of model components across different scales.

The success of our multi-resolution optimization approach provides evidence that model representations are naturally organized hierarchically, aligning with recent work in circuit-style interpretability \citep{elhage2022mathematical, olah2020zoom}. This suggests that models may learn to decompose visual information across multiple scales not just for discrimination tasks, but as a fundamental organizational principle that supports both discriminative and generative capabilities.

Perhaps most intriguingly, the ability to extract high-quality generative behavior from discriminative models challenges the traditional dichotomy between discriminative and generative architectures from an interpretability perspective. This implies that the apparent distinction between these model types may be more a function of how we access their capabilities than of fundamental differences in their learned representations \citep{bau2019gan}.

Beyond its primary application as a synthesis method, DAS can be viewed as a novel interpretability technique. By revealing what generative information is preserved in discriminative models, it provides a new lens for understanding what these models actually learn. This suggests that the space of possible interpretability tools may be much richer than previously recognized, particularly when we consider emergent capabilities that span traditional architectural boundaries.

\subsection{Limitations and Open Questions}
While DAS achieves impressive results, several important challenges remain: Generation quality can vary across runs and prompts, suggesting room for improving optimization stability. While empirically effective, we lack a complete theoretical framework (that is emerging for diffusion models, e.g. \citet{kamb2024analytictheorycreativityconvolutional}) explaining why multi-resolution optimization so effectively prevents adversarial solutions. 

Our results suggest an intriguing connection between adversarial robustness and generative capabilities. The same multi-resolution structure that enables coherent generation also appears to prevent adversarial patterns, suggesting that natural image statistics may play a crucial role in both phenomena. This raises the possibility that advances in understanding one area could inform the other -- perhaps robust models are inherently better at generation, or generative capabilities could be used as a proxy for robustness. (\citet{fort2025noteonimplementation} provides early indications that adaptive adversarial attacks against a strong multi-resolution model often produce human-interpretable changes to the image.)

\subsection{Future Directions}
Our work opens several promising avenues for future research:

\paragraph{Unified Training Objectives.} Future work could explore training objectives that explicitly optimize for both discriminative and generative capabilities, potentially leading to more efficient and versatile models. This might involve novel loss functions that balance feature discrimination with generative consistency.

\paragraph{Theoretical Frameworks.} Developing formal mathematical frameworks that unify discriminative and generative learning could provide deeper insights into why methods like DAS work. This might draw on information theory, optimal transport, or other theoretical tools to characterize the relationship between these traditionally separate paradigms.

\paragraph{Cross-Domain Applications.} The principles underlying DAS might extend beyond vision to other domains where discriminative and generative tasks have traditionally been separated, such as natural language processing or audio synthesis. This could lead to new training-free generation methods across multiple modalities.

\paragraph{DAS + explicit generative training.} Given that DAS can elicit generative capabilities from discriminative models, it would be intriguing to explore whether its generation can be further improved by additional training explicitly geared towards generation. Merging DAS with diffusion models might be such an avenue. 

\paragraph{Using intermediate layers.} In DAS we have been using the final layer embedding to guide the generation process. In line with the self-ensemble approach in \citet{fort2024ensembleeverywheremultiscaleaggregation}, non-final layers could be used as well, providing a more detailed control over the generation process on different levels of abstraction.

\section{Conclusion}
We have presented Direct Ascent Synthesis, demonstrating that high-quality image generation is possible through direct optimization of discriminative model representations. This finding challenges conventional assumptions about the necessity of dedicated generative training and suggests new directions for understanding and advancing visual synthesis. Our work reveals a deep connection between model inversion, adversarial examples, and image generation—problems that have traditionally been studied separately but share fundamental mathematical characteristics.

The key insight of DAS is that the challenges of model inversion, which have primarily been viewed through the lens of adversarial attacks, can be transformed into opportunities for synthesis through careful regularization. Where adversarial attacks exploit the degeneracy of the inversion problem to find perceptually misleading solutions, our multi-resolution approach harnesses this same flexibility to find natural, semantically meaningful images. This suggests that the perceived limitations of discriminative models—their vulnerability to adversarial examples and the apparent difficulty of inverting their representations—may actually reflect unexploited generative capabilities.

The success of our simple approach raises fundamental questions about the nature of visual representation in neural networks. Perhaps the sharp distinction between discriminative and generative models has been somewhat artificial—both types of models may be learning similar underlying representations, just accessed in different ways. The fact that adversarial patterns and natural images can arise from the same optimization process, differentiated only by their scale-space structure, hints at deep connections between robustness, generalization, and generation in neural networks. It appears that seeds of generation are hidden in all discriminative models.

Looking forward, this work suggests that the boundaries between discrimination, generation, and robustness may be more fluid than previously recognized. By viewing these challenges through a unified lens of representation and optimization, we may discover new approaches that simultaneously advance all three areas. DAS represents an important step in this direction, demonstrating that with appropriate optimization techniques, discriminative models can transcend their traditional role and serve as powerful tools for image synthesis.

\newpage
\bibliography{main}
\bibliographystyle{icml2025}

\end{document}